\documentclass[10pt,letterpaper,twocolumn]{article}
\usepackage[T1]{fontenc}
\usepackage{newtxtext}
\usepackage[margin=0.75in,columnsep=0.28in]{geometry}
\usepackage[hyphens]{url}
\usepackage{graphicx}
\usepackage[numbers,sort&compress]{natbib}
\usepackage{caption}
\usepackage{booktabs}
\usepackage{amsmath}
\usepackage{amssymb}
\usepackage{amsfonts}
\usepackage{mathtools}
\usepackage{newtxmath}
\usepackage{bm}
\usepackage{multirow}
\usepackage{makecell}
\usepackage[table]{xcolor}
\usepackage{algorithm}
\usepackage{algorithmic}
\usepackage{float}
\usepackage{placeins}
\usepackage{balance}
\usepackage[hidelinks]{hyperref}
\usepackage{microtype}

\title{\textbf{EvoSpec: Evolving Speculative Decoding via\\Real-Time Vocabulary and Parameter Adaptation}}
\author{%
  \begin{tabular}{ccccc}
    \textbf{Shuyu Zhang} &
    \textbf{Lingfeng Pan} &
    \textbf{Qicheng Wang} &
    \textbf{Yaqi Shi} &
    \textbf{Yueyang Tan}
  \end{tabular}\\[-0.05em]
  \begin{tabular}{cccc}
    \textbf{Ruyu Yan} &
    \textbf{Jiaqi Chen} &
    \textbf{Lixing Du} &
    \textbf{Lu Wang}$^{*}$
  \end{tabular}\\[0.45em]
  \normalsize School of Computer Science and Technology\\
  Xidian University\\
  Xi'an, China\\
  \texttt{wanglu@xidian.edu.cn}
}
\date{}

\hypersetup{
  pdftitle={EvoSpec: Evolving Speculative Decoding via Real-Time Vocabulary and Parameter Adaptation},
  pdfauthor={Shuyu Zhang, Lingfeng Pan, Qicheng Wang, Yaqi Shi, Yueyang Tan, Ruyu Yan, Jiaqi Chen, Lixing Du, Lu Wang}
}

\begin{document}

\maketitle
\begin{abstract}
  Speculative decoding accelerates Large Language Model inference through draft-then-verify generation, yet lightweight draft models face coupled efficiency and quality limitations: large-vocabulary output projection is costly, while limited draft capacity and static parameters reduce acceptance under specialized or shifting inputs. Vocabulary pruning lowers projection cost, but static variants miss locally important long-tail tokens, while dynamic variants remain sensitive to preset selection policies and budgets. Moreover, limited draft capacity can leave the draft distribution misaligned even when the target token is covered. Online alignment improves draft quality, but full-parameter updates introduce substantial memory and latency overhead. We introduce EvoSpec, which jointly adapts the active vocabulary and lightweight draft parameters from verification feedback. EvoSpec asynchronously retrieves semantic and statistical token neighbors and performs curriculum-weighted online LoRA alignment while preserving exact target-model verification. On Qwen3-8B/EAGLE-2, EvoSpec reaches a $2.18\times$ speedup over vanilla decoding and a $1.20\times$ gain over EAGLE-2, while improving specialized-domain coverage and using $27\%$ less auxiliary GPU adaptation memory than full-parameter online adaptation.
\end{abstract}

\section{Introduction}
\label{sec:introduction}

Large Language Models (LLMs) generate tokens autoregressively, making inference latency grow with output length~\cite{kim2023speculative,fu2024break}. Speculative Decoding (SD) reduces this serial cost by asking a lightweight draft model to propose multiple tokens that the target model verifies in parallel~\cite{leviathan2023fast}. Although feature-level methods have made the draft backbone increasingly compact, modern tokenizers for multilingual, code, and domain-specific usage still require projection over vocabularies exceeding 100k tokens~\cite{goel2025vocabtrim}. In this regime, the LM head can account for nearly $50\%$--$60\%$ of draft-generation time~\cite{zhao2025fr}, so reducing projection cost without sacrificing accepted draft length is central to further acceleration.

Vocabulary pruning reduces this cost by restricting draft logits to a smaller output space. Static methods such as FR-Spec and VocabTrim retain a fixed high-frequency vocabulary and preserve exact target-model verification~\cite{zhao2025fr,goel2025vocabtrim}, but a context-independent table can omit rare tokens that become locally probable in conversations, code, or specialized domains~\cite{deng2025drpruning,zhao2025efficientxpert}. Dynamic methods instead construct context-dependent active vocabularies and recover more long-tail tokens~\cite{chen2026nanospec,williams2026speculative}, but their selection policies and budgets are typically fixed before deployment, making coverage sensitive to how well these preset choices match the current inference distribution.

Moreover, this draft--target mismatch can become more pronounced under domain or topic shift: even when the target token is present in the active vocabulary, a lightweight draft model may assign it insufficient probability~\cite{zhou2023distillspec,ganesan2025massv,sun2024learning}. Online adaptation uses verification feedback to improve this alignment~\cite{hong2025training,li2025test}, but full-parameter updates and heavy distillation objectives can add substantial optimizer state, memory traffic, and latency~\cite{liu2023online}. Existing approaches therefore improve either output-space efficiency or draft quality without jointly adapting both under a tight inference budget.

We propose EvoSpec, a closed-loop framework that jointly adapts the active vocabulary and a lightweight parameter path. EvoSpec begins with a compact static core and interprets each verification result as evidence about both vocabulary coverage and draft-target alignment. An out-of-vocabulary failure triggers asynchronous retrieval of semantic neighbors from the LM-head space and statistical neighbors from a sparse co-occurrence graph, with a fixed-budget ARC buffer retaining useful tokens. Covered but misaligned predictions provide supervision for a confidence-aware, curriculum-weighted LoRA update. Both paths operate without changing the target model or its exact verification rule, allowing the draft model to evolve while keeping adaptation off the synchronous verification path.

The main contributions of this paper are summarized as follows:

\begin{itemize}
    \item \textbf{Joint vocabulary and parameter adaptation:} We formulate large-vocabulary SD as a closed-loop process in which target verification identifies both coverage failures and draft-target mismatch, allowing the two error sources to be repaired together.

    \item \textbf{Budgeted online evolution:} Semantic retrieval, sparse-graph expansion, and ARC maintenance recover context-relevant long-tail tokens, while curriculum-driven LoRA aligns the draft distribution with $27\%$ lower auxiliary GPU adaptation memory than full-parameter online updates.

    \item \textbf{End-to-end evaluation:} Experiments on Qwen3-8B/EAGLE-2 show a $2.18\times$ speedup over vanilla decoding and a $1.20\times$ gain over EAGLE-2, with the same trend on Llama-3.2-1B/EAGLE-2. Coverage, ablation, controlled topic-shift, and long-horizon analyses isolate the benefits of the two adaptation paths.
\end{itemize}
\section{Related work}

The autoregressive nature of LLM generation results in memory-bound latency, where inference speed is constrained by memory bandwidth rather than compute power~\cite{xia2024unlocking, yuan2024llm}. SD addresses this by verifying multiple low-cost draft tokens within a single parallel forward pass of the target model~\cite{zhou2024survey}. Early SD methods relied on independent draft models, which are typically smaller versions of the target~\cite{leviathan2023fast, chen2023accelerating}. However, the efficiency of these methods is often constrained by the significant computational and memory overhead required to maintain a separate model architecture, limiting the net acceleration gain in practical settings~\cite{zhang2024draft}. To mitigate this, state-of-the-art approaches have shifted toward feature-level fusion. Medusa~\cite{cai2024medusa} augments the target model with multiple decoding heads by utilizing the target's own high-level representations to predict future tokens. EAGLE series~\cite{li2024eagle, li2024eagle2, li2025eagle} further advance this by introducing a lightweight autoregressive head operating on the target's feature space. Despite reducing draft generation costs, these methods still necessitate a computationally expensive projection over the full vocabulary at the draft head's final layer, which persists as a major latency source~\cite{cheng2024recurrent}.

This vocabulary bottleneck becomes particularly acute as LLMs expand to support multilingual and code generation capabilities, causing vocabulary sizes to surge. Recent analyses, such as FR-Spec~\cite{zhao2025fr} and VocabTrim~\cite{goel2025vocabtrim}, identify the LM Head projection cost as a primary bottleneck for lightweight draft models, accounting for up to $60\%$ of the draft generation time. Existing solutions primarily employ static vocabulary pruning, constructing a context-independent subset of global high-frequency tokens to reduce the search space. However, such methods rely on the assumption of static word frequencies and fail to capture long-tail tokens that become locally probable in specialized domains or during topic shifts~\cite{holtzman2019curious,khandelwal2019generalization}. In SD, these missing long-tail candidates manifest as OOV events and lower acceptance rates under domain shift, as also observed by recent vocabulary-pruned SD methods~\cite{zhao2025fr,goel2025vocabtrim}. Dynamic vocabulary methods show that context-dependent active vocabularies can restore acceptance length~\cite{chen2026nanospec,williams2026speculative}, but their selection policies and budgets remain fixed before deployment and can be sensitive to distribution changes. Vocabulary selection also leaves draft--target miscalibration on covered tokens as a separate bottleneck. EvoSpec builds on this direction by retrieving missing long-tail candidates with HNSW-based approximate nearest-neighbor search on the LM Head weight matrix~\cite{malkov2018efficient}, then using verification feedback to align the draft distribution on the evolving subspace.

Beyond architectural and vocabulary constraints, a critical limitation of standard SD is the distribution shift between the pre-trained draft model and the specific input distribution encountered during inference~\cite{sun2024learning}. Static draft models often perform poorly when processing user-specific prompts or novel topics. Online Speculative Decoding (OSD)~\cite{liu2023online} addresses this by treating verification signals from the target model as supervision to fine-tune the draft model via real-time distillation. Similarly, Test-Time Training paradigms update model parameters on test data to minimize prediction entropy or perplexity~\cite{sun2020test}. However, naive full-parameter updates are computationally expensive and prone to instability. Integrating these insights, EvoSpec employs a lightweight Online LoRA mechanism~\cite{hu2022lora}. This approach differs from the update strategies in OSD and provides a resource-efficient mechanism for adaptation to specialized distributions.

\section{Methodology}
\label{sec:method}

\subsection{Problem formulation}

We formulate the problem within the general context of SD. Given a target model $\mathcal{M}_p$ and a draft model $\mathcal{M}_q$, the objective is to maximize the effective generation speedup. Let $x_{<t}$ denote the input context at step $t$. The draft model generates a sequence of tokens which are verified by $\mathcal{M}_p$ based on the rejection sampling criterion $\alpha$~\cite{leviathan2023fast}.

Feature-level SD methods such as EAGLE-2 provide a strong acceleration backbone~\cite{li2024eagle2}, but they share a critical computational bottleneck with standard approaches: the output projection. For any draft model, predicting the next token requires projecting the hidden state $h_t \in \mathbb{R}^d$ onto the full vocabulary $\mathcal{V}$ via the language model head $W_{LM} \in \mathbb{R}^{|\mathcal{V}| \times d}$:
\begin{equation}
\label{eq:projection}
    p_{\mathcal{M}_q}(x \mid x_{<t}) = \text{Softmax}(h_t W_{LM}^T).
\end{equation}
The complexity $\mathcal{O}(d \times |\mathcal{V}|)$, where $d$ is the hidden dimension, becomes prohibitive when $|\mathcal{V}| > 100\text{k}$, making output-space reduction valuable across lightweight SD architectures~\cite{zhao2025fr}.

We formulate the acceleration of large-vocabulary models as a dynamic resource allocation problem. At each inference step $t$, EvoSpec first constructs a compact local vocabulary subspace $\mathcal{V}_t \subset \mathcal{V}$ under a bounded dynamic budget:
\begin{equation}
\label{eq:optimization}
\begin{aligned}
    \min_{\mathcal{V}_t} \quad & |\mathcal{V}_t| \\
    \text{s.t.} \quad
    & \sum_{x \in \mathcal{V}_t}
    p_{\mathcal{M}_p}(x \mid x_{<t})
    \geq 1 - \epsilon_{\mathrm{cov}}, \\
    & |\mathcal{V}_t \setminus \mathcal{V}_{\text{static}}|
    \leq N_{\mathrm{dyn}}.
\end{aligned}
\end{equation}
where $\epsilon_{\mathrm{cov}}$ is the tolerated probability-mass loss and $N_{\mathrm{dyn}}$ is the dynamic-buffer budget. Since the true target distribution $p_{\mathcal{M}_p}$ is unavailable during the drafting phase, EvoSpec approximates this optimization via retrieval-based subspace construction, while the complementary parameter update $\Delta\theta_t$ is handled by the online alignment objective in Eq.~\ref{eq:lora_objective}.

\subsection{System architecture}

\begin{figure*}[t]
  \centering
  \includegraphics[width=\textwidth]{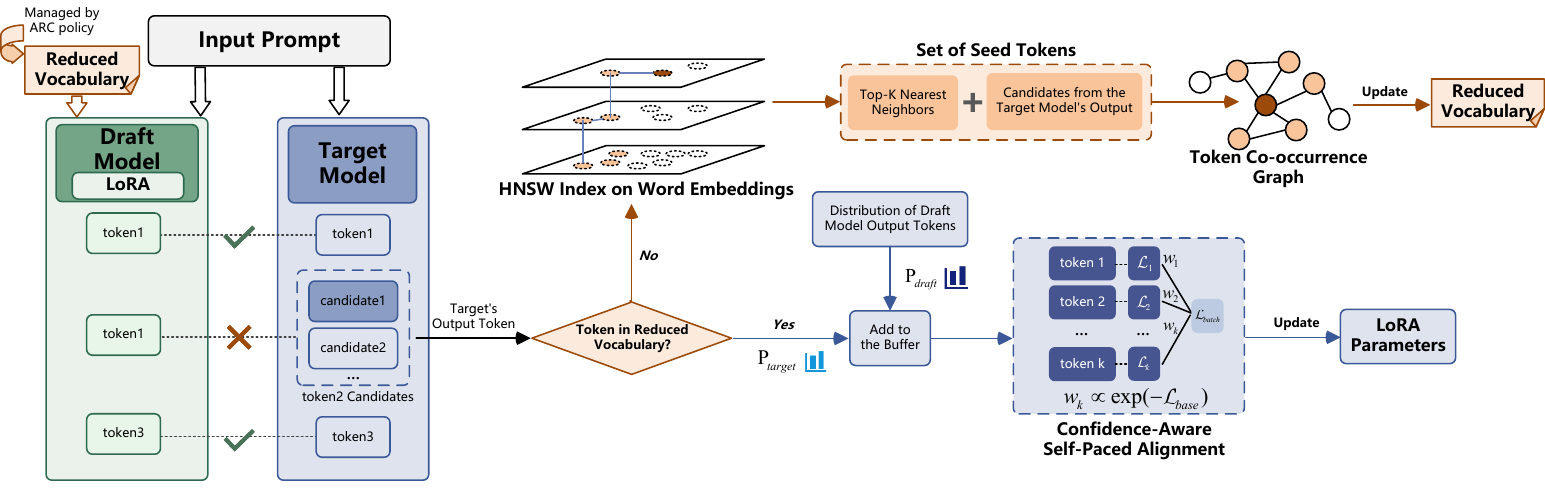}
  \caption{The architecture of EvoSpec. The system employs a decision logic based on vocabulary coverage to trigger two parallel loops: (1) the top branch asynchronously recalls semantic and statistical neighbors on the CPU upon OOV detection; and (2) the bottom branch fine-tunes the draft model's LoRA parameters using a self-paced curriculum for tokens covered by the current vocabulary.}
  \label{fig:framework}
\end{figure*}

To approximate the optimization solution defined in Eq.~\eqref{eq:optimization} under real-time constraints, EvoSpec establishes a closed-loop feedback system as illustrated in Figure~\ref{fig:framework}. The core philosophy is to decouple the intractable joint optimization into two asynchronous processes triggered by the target model's output token status relative to the current reduced vocabulary.

\begin{itemize}
    \item \textbf{Path A: Vocabulary expansion.} When the verified token $x_t$ falls outside the current subspace $\mathcal{V}_t$, EvoSpec treats the event as a practical signal of insufficient local coverage. The system activates the dynamic vocabulary generator to recall missing terms. To mitigate the retrieval latency of Hierarchical Navigable Small World (HNSW), this module operates asynchronously on the CPU. Retrieved candidates are merged into the active buffer when available; if retrieval misses the next draft launch, that launch proceeds without them rather than blocking verification.

    \item \textbf{Path B: Parameter alignment.} When $x_t$ resides within $\mathcal{V}_t$, the sample becomes eligible for online alignment rather than being updated unconditionally. It enters the replay buffer only if the retained-support KL divergence exceeds $\epsilon_{\mathrm{align}}$ or the target rejects the draft proposal; OOV failures are routed exclusively to Path A so that missing support is not treated as parameter error. The confidence-aware curriculum then weights qualified samples so that LoRA updates reduce distributional divergence while preserving training stability.
\end{itemize}

This bifurcated architecture allows EvoSpec to dynamically expand the search space $\mathcal{V}_t$ and adapt parameters $\theta_t$ simultaneously, evolving the draft model to fit the target distribution on-the-fly without placing retrieval on the synchronous verification path.

\subsection{Dynamic search space reduction}
\label{sec:dynamic_vocab}

To address the optimization objective defined in Eq.~\eqref{eq:optimization}, EvoSpec abandons the rigid full vocabulary in favor of a dynamic construction strategy. We approximate the optimal high-probability set by decomposing the search space into global and local components. At each inference step $t$, the active vocabulary subspace $\mathcal{V}_t$ is formulated as the union of a static core and context-dependent dynamic subsets:
\begin{equation}
    \label{eq:vocab_union}
    \mathcal{V}_t = \mathcal{V}_{\text{static}} \cup \mathcal{S}_{\text{sem}}(h_t) \cup \mathcal{S}_{\text{graph}}(\mathcal{S}_{\text{sem}}),
\end{equation}
where $\mathcal{V}_{\text{static}}$ denotes the top-$K$ frequent tokens representing the stationary distribution of general syntax. We establish $\mathcal{S}_{\text{sem}}$ and $\mathcal{S}_{\text{graph}}$ through embedding-based indexing and graph-based expansion, as described below:

\paragraph{Semantic approximation via HNSW.}
The projection over the LM head induces a Maximum Inner Product Search (MIPS) problem. To bypass $O(|\mathcal{V}|)$ exhaustive search, we index the target model's output embeddings $W_{\mathrm{LM}}$ using HNSW with norm augmentation~\cite{malkov2018efficient}. Upon OOV detection after verification, we use the target-side hidden state at the rejected speculative position from the same verification pass to retrieve the top-$N$ nearest neighbors $\mathcal{S}_{\mathrm{sem}}$. This provides sublinear approximate search in practice and captures topic-specific terms missed by static tables.

At each failure event, the initial pool unions the target top-10 tokens with the HNSW top-10 neighbors; graph expansion then contributes up to eight successors per seed before deduplication and ARC admission. Because this feedback arrives after verification, the retrieved neighborhood is used by subsequent draft launches rather than retroactively changing the verified token.

\paragraph{Statistical expansion via sparse graph.} Since semantic similarity often overlooks rigid sequential patterns, we incorporate statistical dependencies via a token co-occurrence graph $\mathcal{G} = (\mathcal{V}, \mathcal{E})$. Constructed from the SlimPajama corpus~\cite{cerebras2023slimpajama}, the graph is pruned by a threshold $\tau$ to ensure sparsity. We query $\mathcal{G}$ to identify high-probability successors for tokens in $\mathcal{S}_{\text{sem}}$. The resulting set $\mathcal{S}_{\text{graph}}$ complements semantic retrieval by capturing fixed collocations and terminological constraints.

\paragraph{Subspace maintenance under budget constraints.} To prevent unbounded expansion, we enforce a fixed dynamic-buffer budget $|\mathcal{V}_{\mathrm{dyn},t}| \leq N_{\mathrm{dyn}}$ (equivalently, $|\mathcal{V}_t| \leq |\mathcal{V}_{\text{static}}| + N_{\mathrm{dyn}}$). We employ an Adaptive Replacement Cache (ARC) policy that dynamically balances recency and frequency to manage token eviction. This ensures $\mathcal{V}_t$ evolves to retain only the most context-relevant tokens, thereby stabilizing inference latency.

\subsection{Test-time adaptation via online alignment}
\label{sec:online_alignment}

While the dynamic vocabulary addresses the recall bottleneck, the static parameters $\theta_{\text{fixed}}$ of the draft model inevitably suffer from distributional shift when facing unseen topics. To mitigate this, we introduce a lightweight online alignment strategy. Instead of indiscriminate updates, we propose a confidence-aware curriculum to stabilize the adaptation process.

\paragraph{Adaptive horizon weighting.} SD inherently involves hierarchical uncertainty: the first token prediction ($j=1$) serves as the anchor, while subsequent predictions ($j>1$) are subject to error accumulation. Naively optimizing all steps with equal weight introduces significant gradient noise from distal, low-confidence tokens. To address this, we construct a dynamic weighting curriculum using the first-step cross-entropy loss against the verified token, $\mathcal{L}_{\text{base}}$, as a real-time proxy for alignment confidence. The loss weight $w_j$ for the $j$-th speculative step is formulated as:
\begin{equation}
    \label{eq:curriculum_weight}
    w_j = \exp\left( - \beta \cdot \mathcal{L}_{\text{base}} \cdot (j-1) \right),
\end{equation}
where $\beta$ controls the decay strength across speculative steps; its effect is examined in the sensitivity analysis. This mechanism establishes an auto-evolving curriculum. Under high uncertainty (large $\mathcal{L}_{\text{base}}$), $w_j$ decays rapidly to filter out noisy long-range gradients, effectively reducing the objective to Next-Token Prediction. Conversely, under low uncertainty (small $\mathcal{L}_{\text{base}}$), the weight decay flattens as the draft model aligns with the context, automatically expanding the effective horizon to capture longer-range dependencies.

\paragraph{Online parameter update.} To enable efficient adaptation, we employ LoRA by freezing the pre-trained draft backbone $\theta_{\text{fixed}}$ and only optimizing the rank-decomposition matrices $\Delta \theta$ injected into the query and value projections ($W_q, W_v$). Instead of standard cross-entropy, we implement distribution-level knowledge distillation. We maintain a buffer $\mathcal{B}$ storing verified trajectories and their corresponding top-$K_{\mathrm{logit}}$ logits $z$ from the target model, with $K_{\mathrm{logit}}=256$. These are converted into a soft target distribution $\hat{p}_{\mathcal{M}_p}$ via a distillation temperature $T_{\mathrm{KD}}$, which is independent of the generation temperature. The optimization objective minimizes the weighted KL divergence over the speculation horizon $\gamma$:
\begin{equation}
    \label{eq:lora_objective}
    \min_{\Delta \theta} \mathcal{J} = \sum_{(x, z) \in \mathcal{B}} \sum_{j=1}^{\gamma} w_j \cdot T_{\mathrm{KD}}^2 \cdot D_{\mathrm{KL}}\left( \hat{p}_{\mathcal{M}_p}^{(j)} \parallel \tilde{p}_{\mathcal{M}_q}^{(j)} \right),
\end{equation}
where $\hat{p}_{\mathcal{M}_p}^{(j)} = \text{Softmax}(z_{\mathcal{M}_p}^{(K_{\mathrm{logit}},j)} / T_{\mathrm{KD}})$ and $\tilde{p}_{\mathcal{M}_q}^{(j)} = \text{Softmax}(z_{\mathcal{M}_q}^{(K_{\mathrm{logit}},j)} / T_{\mathrm{KD}})$ are temperature-scaled distributions on the same retained logit support. This aligns the draft's probability landscape with the target's while avoiding the high memory overhead of storing full vocab logits. We use $\epsilon_{\mathrm{align}}=0.1$ for the admission gate. Eligible samples accumulate until the replay buffer reaches $B=32$; the controller then launches one asynchronous LoRA update and clears the buffer, avoiding per-token optimization.

\section{Empirical evaluation}
\label{sec:experiments}

\subsection{Experimental setup}
\label{sec:setup}

\paragraph{Datasets.}
We evaluate EvoSpec on general, code, and specialized-domain workloads:
\begin{itemize}
    \item \textbf{General workloads:} Spec-Bench~\cite{xia2024unlocking} covers WMT14 (MT, DE-EN)~\cite{bojar-EtAl:2014:W14-33}, MT-Bench (Conversation)~\cite{zheng2023judging}, Natural Questions for RAG and QA~\cite{kwiatkowski2019natural}, GSM8K (Math)~\cite{cobbe2021gsm8k}, and CNN/DM (Summarization)~\cite{nallapati2016abstractive}. We use 80 entries per subtask and a maximum generation length of 1024.
    \item \textbf{Code and specialized domains:} HumanEval~\cite{chen2021evaluating} contains 164 code-generation problems and uses a maximum generation length of 512. We use Pile of Law~\cite{henderson2022pile} and PubMedQA~\cite{jin2019pubmedqa} for legal and medical evaluation. The controlled adaptation analysis concatenates 100 HumanEval requests with 100 requests drawn evenly from the legal and medical workloads.
    \item \textbf{Long-horizon conversations:} LMSYS-Chat-1M~\cite{zheng2023lmsyschat1m} provides real multi-turn conversations. We evaluate 144 English trajectories with more than 50 turns whose replayed histories remain within the native context window under the 256-token response cap.
\end{itemize}

\paragraph{Models.}
The main experiments use Qwen3-8B-Instruct~\cite{qwen3technicalreport}, which has a 151k-token vocabulary, as the target model. The draft model follows EAGLE-2~\cite{li2024eagle2}: a single-layer speculative module that reuses the target embedding and LM head. We also evaluate Llama-3.2-1B-Instruct with EAGLE-2 in Table~\ref{tab:end_to_end_speed}.

\paragraph{Baselines.}
We compare EvoSpec against four distinct categories of methods:
(1) \textbf{Vanilla decoding:} Autoregressive decoding without speculative drafting.
(2) \textbf{Full-vocabulary speculative decoding:} We include EAGLE-2 as the unpruned speculative-decoding baseline.
(3) \textbf{Static and dynamic vocabulary pruning:} FR-Spec uses a 32k static vocabulary derived from SlimPajama corpus statistics~\cite{cerebras2023slimpajama,zhao2025fr}, while SpecVocab~\cite{williams2026speculative} uses its reported 2k context-dependent active vocabulary before draft-logit computation. Both baselines use the same prompts, greedy decoding, EAGLE-2 search depth, and draft-token budget as EvoSpec.
(4) \textbf{Online Adaptation:} We compare against OSD~\cite{liu2023online} using the same 32k active vocabulary and update schedule as the alignment-only configuration, isolating the effect of its full-parameter update path from vocabulary size.

\paragraph{Metrics and implementation details.}
We report Mean Acceptance Length (MAL), end-to-end throughput, speedup, and peak auxiliary GPU memory. Throughput is measured per request and averaged before computing speedups; $\pm$ denotes sample standard deviation over prompts or, for long-horizon evaluation, trajectories. The reported $95\%$ confidence intervals use 10,000 paired bootstrap resamples of the seven task-level relative throughput gains. All methods use greedy decoding, EAGLE-2 search depth 6, and a 60-token draft budget. The default EvoSpec configuration is 32k+D256 with LoRA rank 32, learning rate $10^{-5}$, and $\beta=0.3$. Qwen3-8B experiments use BF16 weights on dual NVIDIA GeForce RTX 4090 GPUs (24GB each). End-to-end timing includes vocabulary maintenance and asynchronous alignment with replay buffer $B=32$.

\subsection{Draft quality and inference speedup}
\label{sec:main_results}

We evaluate EvoSpec on top of EAGLE-2 in the large-vocabulary setting used by FR-Spec, examining whether context-aware retrieval and online alignment can recover acceptance length while preserving end-to-end acceleration.

\paragraph{Validating dynamic expansion and its vocabulary budget.} First, we verify whether our dynamic vocabulary mechanism mitigates the acceptance-length degradation caused by static pruning and examine how this recovery translates into throughput. Table~\ref{tab:main_perf} jointly reports average MAL, throughput, and speedup across Spec-Bench plus HumanEval under the Qwen3-8B/EAGLE-2 configuration.

The static baseline FR-Spec exhibits a clear regression, with average MAL dropping from 3.74 to 3.57. SpecVocab recovers a large portion of this loss with a context-conditioned vocabulary, reaching 3.69 on average. For EvoSpec, an 8k static base is too aggressive for a small dynamic buffer, while 16k becomes competitive and 32k retains enough common-token mass for the dynamic budget to focus on locally relevant long-tail tokens. With a 32k base, the 256-candidate setting reaches 3.72 MAL ($99.5\%$ of full vocabulary) at 120.17 tokens/s. Increasing the dynamic budget to 512 raises MAL only to 3.73 while lowering throughput to 119.69 tokens/s; increasing it to 1024 adds no further MAL gain. We therefore select 32k+256 as the default speed-quality operating point. Section~\ref{sec:coverage} directly evaluates active-vocabulary probability mass and top-$k$ recall at a 256-candidate budget.

\begin{table}[htpb]
    \centering
    \small
    \setlength{\tabcolsep}{0pt}
    \renewcommand{\arraystretch}{1.03}
    \begin{tabular*}{\columnwidth}{@{\extracolsep{\fill}}lccccc@{}}
        \toprule
        \textbf{Method} & \textbf{Static} & \textbf{Dyn.} & \textbf{MAL} & \textbf{tok/s} & \textbf{Speedup} \\
        \midrule
        \textbf{EAGLE-2} & Full & -- & 3.74 & 100.22 & $1.81\times$ \\
        \textbf{FR-Spec} & 32k & -- & 3.57 & 111.75 & $2.02\times$ \\
        \textbf{SpecVocab} & -- & 2k & 3.69 & 116.83 & $2.12\times$ \\
        \midrule
        \multirow{9}{*}{\textbf{EvoSpec}}
        & \multirow{3}{*}{8k} & 256  & 3.35 & 109.99 & $1.99\times$ \\
        & & 512  & 3.42 & 110.87 & $2.01\times$ \\
        & & 1024 & 3.46 & 110.05 & $1.99\times$ \\
        \cmidrule(lr){2-6}
        & \multirow{3}{*}{16k} & 256  & 3.63 & 117.38 & $2.13\times$ \\
        & & 512  & 3.67 & 117.90 & $2.13\times$ \\
        & & 1024 & 3.69 & 116.67 & $2.11\times$ \\
        \cmidrule(lr){2-6}
        & \multirow{3}{*}{32k} & 256  & 3.72 & \textbf{120.17} & \textbf{$2.18\times$} \\
        & & 512  & \textbf{3.73} & 119.69 & $2.17\times$ \\
        & & 1024 & 3.73 & 118.37 & $2.14\times$ \\
        \bottomrule
    \end{tabular*}
    \caption{Draft quality and vocabulary-budget trade-off on Qwen3-8B/EAGLE-2. Static and Dyn. denote the always-active static vocabulary and the context-dependent dynamic budget. Speedup is relative to vanilla autoregressive decoding.}
    \label{tab:main_perf}
\end{table}

\paragraph{Speedup and hyperparameter trade-off.} Table~\ref{tab:end_to_end_speed} reports end-to-end throughput for vanilla autoregressive decoding, EAGLE-2, FR-Spec, SpecVocab, and EvoSpec on Qwen3-8B and Llama-3.2-1B.

FR-Spec improves the average throughput from 100.22 to 111.75 tokens/s, a $1.12\times$ gain over EAGLE-2. However, this gain comes with lower MAL, so part of the projection saving is spent on more frequent verification failures. SpecVocab reaches 116.83 tokens/s, while EvoSpec (Static 32k + Dynamic 256) reaches 120.17 tokens/s, corresponding to $2.18\times$ over vanilla decoding and $1.20\times$ over EAGLE-2. Its mean throughput gain over SpecVocab is $2.9\%$ ($95\%$ paired bootstrap CI: $0.5\%$--$5.2\%$), computed over the seven task-level relative gains. We adopt \textbf{Static 32k + Dynamic 256} as the default because the 512-token buffer provides only marginal MAL recovery while adding slightly more maintenance overhead.

The Llama-3.2-1B block preserves the same ordering: EAGLE-2, FR-Spec, SpecVocab, and EvoSpec (Dynamic 256) reach 323.51, 390.13, 412.31, and 422.59 tokens/s, respectively. EvoSpec corresponds to $1.68\times$ over vanilla decoding and $1.31\times$ over EAGLE-2, with a $2.5\%$ mean gain over SpecVocab ($95\%$ paired bootstrap CI: $2.3\%$--$2.8\%$).

\begin{table*}[t]
    \centering
    \small
    \setlength{\tabcolsep}{2.0pt}
    \renewcommand{\arraystretch}{1.05}
    \begin{tabular*}{\textwidth}{@{\extracolsep{\fill}}clccccccccc@{}}
        \toprule
        \textbf{Model} & \textbf{Config.} & \textbf{MT} & \textbf{Conv.} & \textbf{RAG} & \textbf{Math} & \textbf{QA} & \textbf{Summ.} & \textbf{Code} & \textbf{Avg.} & \textbf{Speedup} \\
        \midrule

        \multirow{6}{*}{\textbf{Qwen3-8B}}
        & Vanilla
        & 56.46 & 56.11 & 51.84 & 56.81 & 56.37 & 53.76 & 55.28
        & 55.23
        & $1.00\times$ \\
        
        & EAGLE-2
        & 97.82 & 112.63 & 92.96 & 116.23 & 92.65 & 92.32 & 96.93
        & 100.22
        & $1.81\times$ \\
        
        & FR-Spec (32k)
        & 111.22 & 125.79 & 102.19 & 131.68 & 105.19 & 104.10 & 102.07
        & 111.75
        & $2.02\times$ \\

        & SpecVocab
        & 115.10 & 131.80 & 107.80 & 138.00 & 108.40 & 103.80 & 112.91
        & 116.83
        & $2.12\times$ \\

        & \textbf{EvoSpec (32k+D256)}
        & \textbf{117.73} & \textbf{135.55} & \textbf{110.63} & \textbf{141.83} & \textbf{113.62} & \textbf{112.35} & \textbf{109.46}
        & \textbf{120.17}
        & \textbf{$2.18\times$} \\

        & EvoSpec (32k+D512)
        & 117.44 & 134.97 & 110.49 & 141.24 & 113.27 & 111.87 & 108.58
        & 119.69
        & $2.17\times$ \\

        \midrule

        \multirow{6}{*}{\textbf{Llama-3.2-1B}}
        & Vanilla
        & 259.83 & 255.89 & 220.25 & 263.34 & 260.13 & 248.15 & 256.64
        & 252.03
        & $1.00\times$ \\

        & EAGLE-2
        & 306.04 & 358.37 & 266.84 & 372.37 & 305.52 & 294.82 & 360.60
        & 323.51
        & $1.28\times$ \\

        & FR-Spec (32k)
        & 378.90 & 428.75 & 317.68 & 467.53 & 378.39 & 363.70 & 395.95
        & 390.13
        & $1.55\times$ \\

        & SpecVocab
        & 398.58 & 455.67 & 336.47 & 494.37 & 401.32 & 386.19 & 413.56
        & 412.31
        & $1.64\times$ \\

        & \textbf{EvoSpec (32k+D256)}
        & \textbf{407.81} & \textbf{466.20} & \textbf{347.10} & \textbf{505.70} & \textbf{411.40} & \textbf{396.30} & \textbf{423.60}
        & \textbf{422.59}
        & \textbf{$1.68\times$} \\

        & EvoSpec (32k+D512)
        & 404.88 & 462.73 & 344.91 & 501.65 & 408.76 & 392.90 & 419.54
        & 419.34
        & $1.66\times$ \\

        \bottomrule
    \end{tabular*}
    \caption{End-to-end speedup on Qwen3-8B and Llama-3.2-1B with EAGLE-2. Speed is reported in tokens/s on the same evaluation datasets. Speedup is computed relative to vanilla decoding within each model block.}
    \label{tab:end_to_end_speed}
\end{table*}

\subsection{Adaptation under a controlled topic shift}
\label{sec:dynamic_adapt}

We construct a 200-request stream with an abrupt distribution shift. The first 100 requests come from HumanEval, and the remaining 100 are drawn evenly from Pile of Law and PubMedQA. This direct concatenation fixes the transition at Request~100 and avoids introducing an auxiliary topic detector into the evaluation. Each plotted point is the request-level MAL in chronological order, without moving-average smoothing.

\begin{figure}[htpb]
    \centering
    \includegraphics[width=\columnwidth]{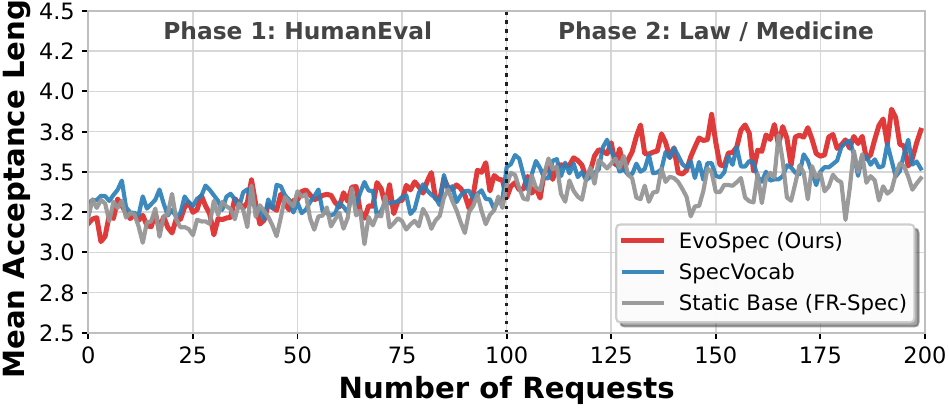}
    \caption{\textbf{Controlled topic-shift adaptation.} The dotted line marks the HumanEval-to-Law/Medicine transition. EvoSpec recovers through online parameter updates, while the baselines retain fixed draft parameters.}
    \label{fig:adaptation_curve}
\end{figure}

Figure~\ref{fig:adaptation_curve} shows that FR-Spec and SpecVocab settle near fixed domain-dependent levels because their draft parameters remain unchanged across requests. EvoSpec experiences a transient drop at the transition, then recovers through verification-driven online alignment and surpasses SpecVocab after approximately 16 requests. Its continued improvement separates parameter adaptation from the coverage benefit of a context-dependent vocabulary.

\FloatBarrier
\subsection{Long-horizon stability on real conversations}
\label{sec:long_horizon}

Beyond the controlled transition, we test adaptation stability on naturally evolving conversations. We replay each selected LMSYS-Chat-1M trajectory chronologically, retain accumulated model-generated context, and cap each response at 256 tokens; no topic labels or event alignment are used. Dynamic vocabulary, replay-buffer, and LoRA states are reset before each trajectory and then evolve only within that conversation, preventing cross-trajectory transfer. Table~\ref{tab:long_horizon} reports trajectory-level mean $\pm$ standard deviation. Although accumulated context lowers absolute throughput, EvoSpec reaches 3.67 MAL and 110.43 tokens/s, improving throughput by $4.1\%$ over SpecVocab and $9.5\%$ over FR-Spec without increasing across-trajectory variability.

\begin{table}[htpb]
    \centering
    \small
    \setlength{\tabcolsep}{0pt}
    \renewcommand{\arraystretch}{1.05}
    \begin{tabular*}{\columnwidth}{@{\extracolsep{\fill}}lcc@{}}
        \toprule
        \textbf{Method} & \textbf{Overall MAL} & \textbf{Throughput (tok/s)} \\
        \midrule
        \textbf{FR-Spec (Static 32k)} & $3.41 \pm 0.06$ & $100.86 \pm 2.37$ \\
        \textbf{SpecVocab} & $3.58 \pm 0.05$ & $106.12 \pm 2.08$ \\
        \textbf{EvoSpec} & $\mathbf{3.67 \pm 0.04}$ & $\mathbf{110.43 \pm 1.86}$ \\
        \bottomrule
    \end{tabular*}
    \caption{Long-horizon LMSYS-Chat-1M evaluation. Results are mean $\pm$ standard deviation across 144 trajectories exceeding 50 turns.}
    \label{tab:long_horizon}
\end{table}

\subsection{Sensitivity analysis of curriculum decay}
\label{sec:sensitivity}

The curriculum decay factor $\beta$ controls the trade-off between filtering noisy long-range supervision and preserving useful multi-token alignment signals. We evaluate $\beta \in \{0.0, 0.1, 0.3, 0.5, 0.7\}$ while keeping the learning rate, optimizer, and update budget fixed. Here $\beta=0.0$ corresponds to an unweighted full-horizon objective rather than a separately retuned baseline.

\begin{figure}[htpb]
    \centering
    \includegraphics[width=\columnwidth]{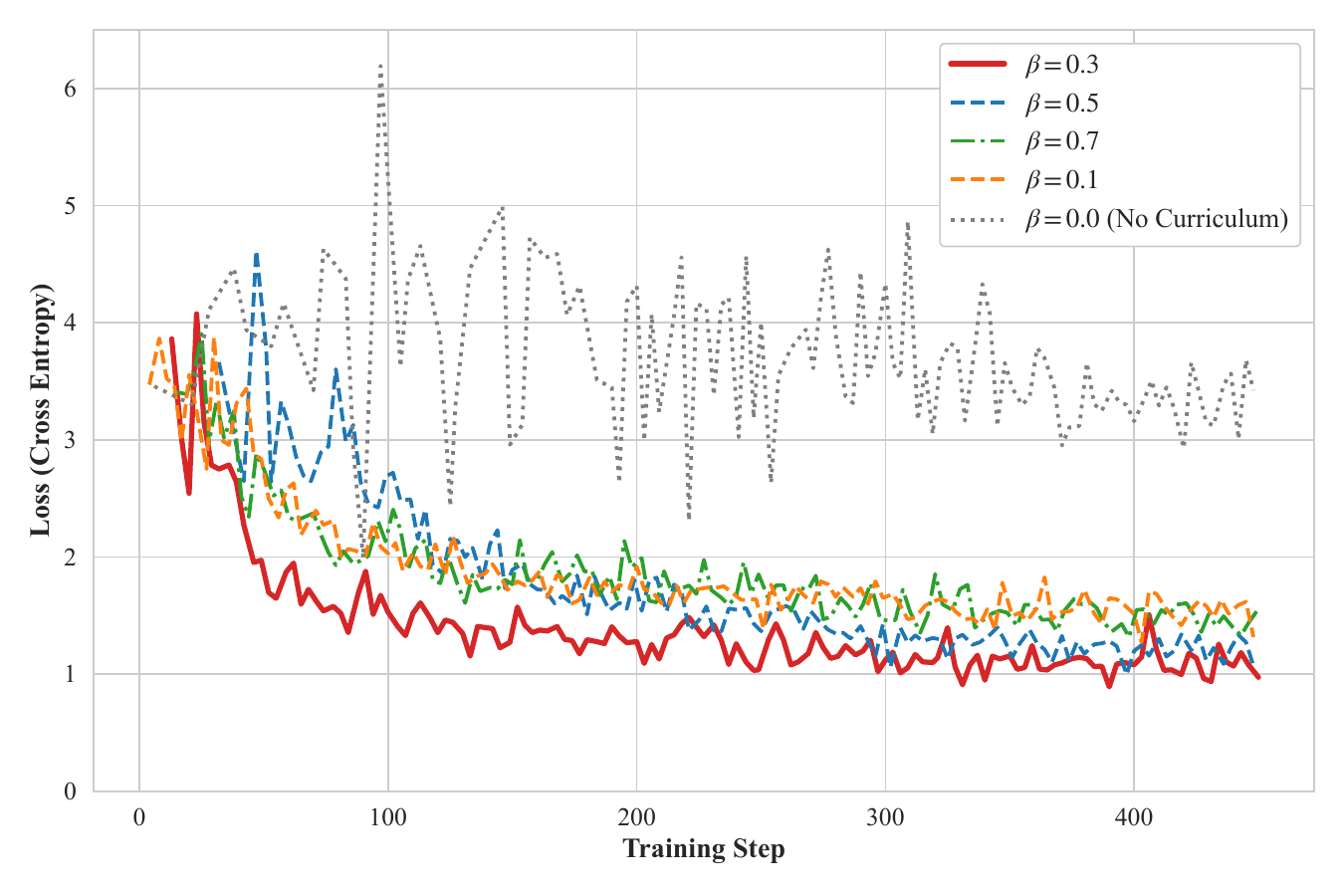}
    \caption{Effect of sensitivity coefficient $\beta$ on convergence. $\beta=0.3$ converges most stably; $\beta=0.0$ fails, whereas larger values over-filter multi-token supervision.}
    \label{fig:impact_beta}
\end{figure}

Figure~\ref{fig:impact_beta} shows that an intermediate decay value gives the most stable convergence. Without curriculum weighting, the loss remains high and oscillatory because unreliable distant targets are overemphasized. Increasing $\beta$ from 0.1 to 0.3 suppresses this noise while retaining sufficient multi-token supervision; larger values over-filter useful trajectories and lead to a higher final loss. We therefore use $\beta=0.3$ in all main experiments. The complementary static-vocabulary and dynamic-budget trade-off is reported in Table~\ref{tab:main_perf}.

\subsection{Ablation study}
\label{sec:ablation}

We evaluate EvoSpec on the Qwen3-8B/EAGLE-2 setup using matched standalone comparisons. Static + OSD and Static + CL-LoRA isolate the parameter-update path under the same 32k vocabulary, while SpecVocab and Dynamic Vocab compare context-dependent vocabulary selection without online parameter updates. Table~\ref{tab:ablation_summary} reports averages across the evaluation datasets together with the complete system.

Each standalone module improves on its corresponding prior approach. CL-LoRA reaches 3.64 MAL and 116.05 tokens/s, compared with 3.62 and 114.02 for OSD, while reducing auxiliary memory from 4,128 to 2,936 MiB. The vocabulary-only configuration reaches 3.70 MAL and 118.45 tokens/s, exceeding SpecVocab at 3.69 and 116.83. The complete system performs best at 3.72 MAL and 120.17 tokens/s. Relative to Static 32k, the standalone throughput gains are 4.30 and 6.70 tokens/s, whereas their combination gains 8.42 tokens/s, or $76.5\%$ of their simple sum. This sub-additivity is expected because the two modules repair partially overlapping draft failures.

\begin{table}[htpb]
    \centering
    \small
    \setlength{\tabcolsep}{0pt}
    \begin{tabular*}{\columnwidth}{@{\extracolsep{\fill}}lcccc@{}}
        \toprule
        \textbf{Method}
        & \makecell{\textbf{Avg.}\\\textbf{MAL}}
        & \textbf{Throughput}
        & \textbf{Speedup}
        & \makecell{\textbf{Aux.}\\\textbf{Mem.}} \\
        \midrule
        \textbf{Static 32k} & 3.57 & 111.75 & $2.02\times$ & 2,641 \\
        \textbf{Static + OSD} & 3.62 & 114.02 & $2.06\times$ & 4,128 \\
        \textbf{Static + CL-LoRA} & 3.64 & 116.05 & $2.10\times$ & 2,936 \\
        \midrule
        \textbf{SpecVocab} & 3.69 & 116.83 & $2.12\times$ & 2,704 \\
        \textbf{Dynamic Vocab} & 3.70 & 118.45 & $2.14\times$ & 2,771 \\
        \midrule
        \textbf{EvoSpec Full} & \textbf{3.72} & \textbf{120.17}
        & \textbf{$2.18\times$} & 3,013 \\
        \bottomrule
    \end{tabular*}
    \caption{Matched module comparison on Qwen3-8B/EAGLE-2. OSD and CL-LoRA share the static vocabulary; both vocabulary-only rows disable parameter updates. Speedup is relative to vanilla decoding and auxiliary memory is in MiB.}
    \label{tab:ablation_summary}
\end{table}

\subsection{Active-vocabulary coverage}
\label{sec:coverage}

Table~\ref{tab:coverage} tests whether throughput gains reflect better coverage rather than projection savings alone. It reports retained target probability mass and top-$k$ recall within each method's 256 highest-ranked context-dependent candidates on Code, Law, and Medicine. FR-Spec is the static reference, while the HNSW and graph rows isolate EvoSpec's retrieval stages.

\begin{table}[htpb]
    \centering
    \small
    \setlength{\tabcolsep}{0pt}
    \begin{tabular*}{\columnwidth}{@{\extracolsep{\fill}}lcccc@{}}
        \toprule
        \textbf{Method} & \textbf{Mass} & \textbf{R@10} & \textbf{R@50} & \textbf{R@100} \\
        \midrule
        \textbf{FR-Spec (32k)}       & 94.2\% & --    & --    & --    \\
        \textbf{SpecVocab (2k)}      & 97.1\% & 93.4\% & 88.4\% & 85.6\% \\
        \textbf{EvoSpec: HNSW}       & 96.1\% & 91.3\% & 85.9\% & 82.6\% \\
        \textbf{EvoSpec: HNSW+Graph} & 97.0\% & 93.2\% & 88.2\% & 85.3\% \\
        \textbf{EvoSpec Vocab Only} & \textbf{97.4\%} & \textbf{93.7\%} & \textbf{88.8\%} & \textbf{86.1\%} \\
        \bottomrule
    \end{tabular*}
    \caption{Active-vocabulary coverage averaged over Code, Law, and Medicine. Mass uses the full active vocabulary; R@$k$ measures target top-$k$ recovery among 256 ranked candidates.}
    \label{tab:coverage}
\end{table}

With a 2k active vocabulary, SpecVocab recovers most target probability mass and slightly exceeds bare HNSW+Graph retrieval. Verification-driven buffer retention yields the complete vocabulary-only module, attaining the highest mass and recall at every cutoff without parameter updates. The full system's remaining gain therefore reflects complementarity with draft-parameter alignment.

\subsection{Resource efficiency analysis}
\label{sec:resource}

Auxiliary GPU memory constrains online SD. CL-LoRA improves MAL and throughput over OSD while cutting alignment memory by $29\%$ (4,128 to 2,936 MiB); the complete system uses 3,013 MiB, \textbf{$27\%$} below full-parameter OSD (Table~\ref{tab:memory_breakdown}). LoRA updates remain asynchronous and amortize to below $0.05$ ms/token. CPU retrieval completes in 2.46 ms, leaving 2.71 ms before the concurrent 5.17 ms Transformer body reaches the reduced-head deadline; late results merge into a subsequent launch without a synchronization stall.

\begin{table}[htpb]
    \centering
    \small
    \setlength{\tabcolsep}{0pt}
    \begin{tabular*}{\columnwidth}{@{\extracolsep{\fill}}lcccc@{}}
        \toprule
        \textbf{Dataset} & 
        \makecell{\textbf{OSD Aux.}\\\textbf{Mem.}} & 
        \makecell{\textbf{EvoSpec Aux.}\\\textbf{Mem.}} & 
        \makecell{\textbf{Update}\\\textbf{Time}} & 
        \makecell{\textbf{Amort.}\\\textbf{Overhead}} \\
        \midrule
        Code & 4,216 & 3,078 & 7.8 ms & 0.043 ms/tok. \\
        Law  & 4,045 & 2,953 & 8.1 ms & 0.047 ms/tok. \\
        Med  & 4,122 & 3,009 & 8.0 ms & 0.045 ms/tok. \\
        \midrule
        \textbf{Avg.} & \textbf{4,128} & \textbf{3,013} & \textbf{8.0 ms} & \textbf{0.045 ms/tok.} \\
        \bottomrule
    \end{tabular*}
    \caption{Resource efficiency across domains. Auxiliary GPU memory excludes frozen target weights, the target-side KV cache, and CPU indices; update overhead is amortized.}
    \label{tab:memory_breakdown}
\end{table}

\section{Conclusion}
\label{sec:conclusion}

EvoSpec uses target verification to jointly adapt the active vocabulary and draft parameters. Asynchronous semantic--statistical retrieval repairs coverage failures, while curriculum-weighted LoRA improves draft--target alignment without changing exact verification. On Qwen3-8B/EAGLE-2, EvoSpec achieves a $2.18\times$ speedup over vanilla decoding with $27\%$ less auxiliary adaptation memory than full-parameter OSD; Llama-3.2-1B shows the same ordering. Ablation and coverage confirm complementary module gains, while controlled-shift and long-horizon evaluations show sustained adaptation under changing and extended contexts.

\section{Limitations}
\label{sec:limitations}

EvoSpec relies on CPU retrieval, offline indices, and sufficient CPU--GPU overlap; limited overlap may reduce latency gains, although asynchronous retrieval does not block verification. Evaluation covers two model families, selected domains, and 144 English trajectories exceeding 50 turns, but not multilingual or substantially longer conversations.

\FloatBarrier
\balance
\bibliographystyle{plainnat}
\bibliography{references}

\clearpage
\onecolumn
\appendix
\section{Implementation Details and Pseudocode}
\label{app:implementation}

\subsection{Algorithm Pseudocode}

Algorithm~\ref{alg:evospec} outlines the EvoSpec inference process. Vocabulary expansion runs asynchronously on the CPU. Event-triggered alignment modifies only the draft model's LoRA adapters. Target verification continues to determine the final output.

\begin{algorithm}[H]
   \caption{EvoSpec Inference Process}
   \label{alg:evospec}
   \small
   \renewcommand{\baselinestretch}{0.92}\selectfont
\begin{algorithmic}[1]
   \STATE \textbf{Input:} Target Model $\mathcal{M}_p$, Draft Model $\mathcal{M}_q(\theta)$, Prompt $x_{<t}$
   \STATE \textbf{Initialize:} Static Vocab $\mathcal{V}_{\mathrm{static}}$, Dynamic Buffer $\mathcal{V}_{\mathrm{dyn}} \leftarrow \emptyset$
   \STATE \textbf{Initialize:} Replay Buffer $\mathcal{B} \leftarrow \emptyset$, Buffer Size $B$
   \STATE \textbf{Hyperparams:} Horizon $\gamma$, Coverage tolerance $\epsilon_{\mathrm{cov}}$, Alignment gate $\epsilon_{\mathrm{align}}$
   \WHILE{not end of generation}
       \STATE $\mathcal{V}_t \leftarrow \mathcal{V}_{\mathrm{static}} \cup \mathcal{V}_{\mathrm{dyn}}$ \COMMENT{Construct active subspace}

       \STATE \textcolor{blue}{\textit{// Draft Phase (GPU)}}
       \STATE Generate draft tree/chain $\hat{x}_{t:t+\gamma}$ using $\mathcal{M}_q$ restricted to $\mathcal{V}_t$

       \STATE \textcolor{blue}{\textit{// Verify Phase (GPU)}}
       \STATE Verify candidates with $\mathcal{M}_p$; obtain accepted tokens, rejected proposals, and target logits
       \STATE Output accepted tokens; update context $x_{<t}$

       \FOR{each verification event $e_i$}
           \STATE $x_i \leftarrow \textsc{TargetToken}(e_i)$; $z_i \leftarrow \textsc{TargetLogits}(e_i)$
           \STATE \textcolor{gray}{\textit{// Path A: Vocabulary Expansion (CPU Async)}}
           \IF{$x_i \notin \mathcal{V}_t$}
               \STATE \textbf{Trigger Async:} \textsc{UpdateVocab}($x_i$, $h_i$)
               \STATE \quad $\hookrightarrow$ HNSW Query: $\mathcal{S}_{\mathrm{sem}} \leftarrow \textsc{Search}(h_i)$
               \STATE \quad $\hookrightarrow$ Graph Query: $\mathcal{S}_{\mathrm{graph}} \leftarrow \textsc{GetNeighbors}(\mathcal{S}_{\mathrm{sem}})$
               \STATE \quad $\hookrightarrow$ Update $\mathcal{V}_{\mathrm{dyn}}$ via ARC policy
           \ENDIF

           \STATE \textcolor{gray}{\textit{// Path B: Parameter Alignment (GPU)}}
           \IF{$x_i \in \mathcal{V}_t$}
               \STATE $d_i \leftarrow D_{\mathrm{KL}}(\hat{p}_{\mathcal{M}_p}(\cdot|x_{<i}) \| \tilde{p}_{\mathcal{M}_q}(\cdot|x_{<i}))$
               \IF{$d_i > \epsilon_{\mathrm{align}}$ \OR \textsc{DraftRejectedByTarget}$(e_i)$}
                   \STATE Add $(x_{<i}, z_i)$ to Replay Buffer $\mathcal{B}$
                   \IF{$|\mathcal{B}| \ge B$}
                       \STATE $\mathcal{L} \leftarrow \textsc{ComputeCurriculumLoss}(\mathcal{B})$
                       \STATE $\theta \leftarrow \theta - \eta \nabla_\theta \mathcal{L}$ \COMMENT{LoRA update}
                       \STATE Clear $\mathcal{B}$
                   \ENDIF
               \ENDIF
           \ENDIF
       \ENDFOR
   \ENDWHILE
\end{algorithmic}
\end{algorithm}

\section{Hyperparameter Settings}

Table~\ref{tab:hyperparams} lists the configuration shared across the experiments unless otherwise specified.

\begin{table}[H]
    \centering
    \small
    \begin{tabular}{lcl}
        \toprule
        \textbf{Category} & \textbf{Parameter} & \textbf{Value} \\
        \midrule
        \multirow{4}{*}{Speculation}
         & Horizon ($\gamma$) & 6 \\
         & Token Budget & 60 \\
         & Temperature & 0.0 (Greedy) \\
         & Top-$p$ & 1.0 \\
        \midrule
        \multirow{8}{*}{Vocabulary}
         & Static Size ($|\mathcal{V}_{\mathrm{static}}|$) & 32,000 \\
         & Dynamic Size ($|\mathcal{V}_{\mathrm{dyn}}|$) & 256 \\
         & HNSW $M$ & 32 \\
         & Graph Pruning $\tau$ & $10^{-4}$ \\
         & Min Co-occurrence Count & 5 \\
         & Max Graph Out-Degree & 64 \\
         & Initial Candidate Set & target top-10 $\cup$ HNSW top-10 \\
         & Graph Expansion per Seed & top-8 outgoing neighbors \\
        \midrule
        \multirow{13}{*}{Online Learning}
         & Buffer Size ($B$) & 32 \\
         & Learning Rate ($\eta$) & $1 \times 10^{-5}$ \\
         & Distillation Temperature ($T_{\mathrm{KD}}$) & 1.0 \\
         & Retained Logits ($K_{\mathrm{logit}}$) & 256 \\
         & Alignment Gate ($\epsilon_{\mathrm{align}}$) & 0.1 \\
         & LoRA Rank ($r$) & 32 \\
         & LoRA Alpha ($\alpha$) & 32 \\
         & Curriculum $\beta$ & 0.3 \\
         & Initial ARC Target ($p$) & 128 \\
         & Ghost-List Capacities & $|B_1|=256,\ |B_2|=256$ \\
         & Max Insertions / Failure Event & 32 \\
         & Min Residency before Eviction & 8 decoding steps \\
         & ARC Warm-up & first 50 events \\
        \midrule
        \multirow{3}{*}{Hardware}
         & GPU & 2 $\times$ RTX 4090 (24GB) \\
         & Precision & BF16 \\
         & CPU & Intel Xeon Platinum \\
        \bottomrule
    \end{tabular}
    \caption{Hyperparameter settings. LoRA parameters are applied only to the draft model's query and value projections ($W_q, W_v$).}
    \label{tab:hyperparams}
\end{table}

\paragraph{Matched component configurations.}
Unless otherwise specified, the default configuration is Qwen3-8B/EAGLE-2 with a 32k static vocabulary and a 256-token dynamic buffer. For the component comparison in the main paper, Static 32k disables both adaptation paths. Static + OSD and Static + CL-LoRA use the same fixed 32k vocabulary, replay-buffer size, and update cadence; they differ only in full-parameter versus LoRA updates. SpecVocab and Dynamic Vocab disable online parameter updates, while EvoSpec Full enables both the dynamic-vocabulary and CL-LoRA paths. Consequently, neither standalone row inherits gains from the other EvoSpec component.

\paragraph{Long-horizon protocol.}
We select 144 English LMSYS-Chat-1M trajectories containing more than 50 turns whose replayed histories remain within the native context window under the 256-token response cap. User turns are replayed in their original order, and prior model-generated turns remain in context. Dynamic vocabulary, replay-buffer, and LoRA states are reset before each trajectory and evolve only within that conversation. MAL and throughput are first averaged within each trajectory and then reported as the mean and standard deviation across trajectories.

\section{Additional Runtime Details}

\paragraph{HNSW semantic retrieval.}
We build the HNSW index offline over the target model's LM-head output embeddings, where each token corresponds to one row of the output projection matrix. We apply norm augmentation so that maximum inner-product retrieval can be implemented with approximate nearest-neighbor search. At runtime, when the verified token falls outside the current reduced vocabulary, the corresponding target-side hidden state from the verification pass is used as the query. HNSW returns the top-$N$ semantic neighbors as the seed set, and the retrieved token IDs are inserted into the active draft vocabulary.

\paragraph{Token co-occurrence graph.}
The token co-occurrence graph is built offline from tokenized SlimPajama using bigram statistics. An edge $u \rightarrow v$ is created whenever token $v$ immediately follows token $u$, with weight $p(v \mid u)$. We apply a minimum co-occurrence count of 5, probability threshold $\tau=10^{-4}$, and maximum out-degree of 64 per source token.

\paragraph{Runtime candidate formation.}
The initial candidate set is the union of the target-side top-10 candidates at the rejection step and the HNSW top-10 semantic neighbors. For each seed token, we retrieve up to eight outgoing graph neighbors, merge the results, and deduplicate them before insertion into the dynamic vocabulary.

\paragraph{ARC-based dynamic vocabulary maintenance.}
The dynamic vocabulary uses Adaptive Replacement Cache (ARC) with capacity 256. ARC maintains recent and frequent lists $T_1$ and $T_2$, together with ghost lists $B_1$ and $B_2$. The target parameter starts at $p=128$ and is constrained to $[0,256]$, with $|B_1|=|B_2|=256$. New tokens enter $T_1$, a second hit promotes them to $T_2$, and repeated hits in $T_2$ move them to the most-recently-used position. When capacity is exceeded, ARC evicts from $T_1$ if $|T_1|>p$ and otherwise from $T_2$.

\paragraph{Additional safeguards.}
We enable batch deduplication, cap insertions at 32 tokens per failure event, require a minimum residency of eight decoding steps before eviction, and keep the ARC target fixed during the first 50 events. Tokens inside the active vocabulary are eligible for alignment only when genuine draft-target mismatch signals are present and the replay buffer is full. The update time is therefore an amortized rather than per-token cost.

\section{Retrieval Latency and Asynchronous Execution}

\begin{figure}[H]
    \centering
    \includegraphics[width=0.95\textwidth]{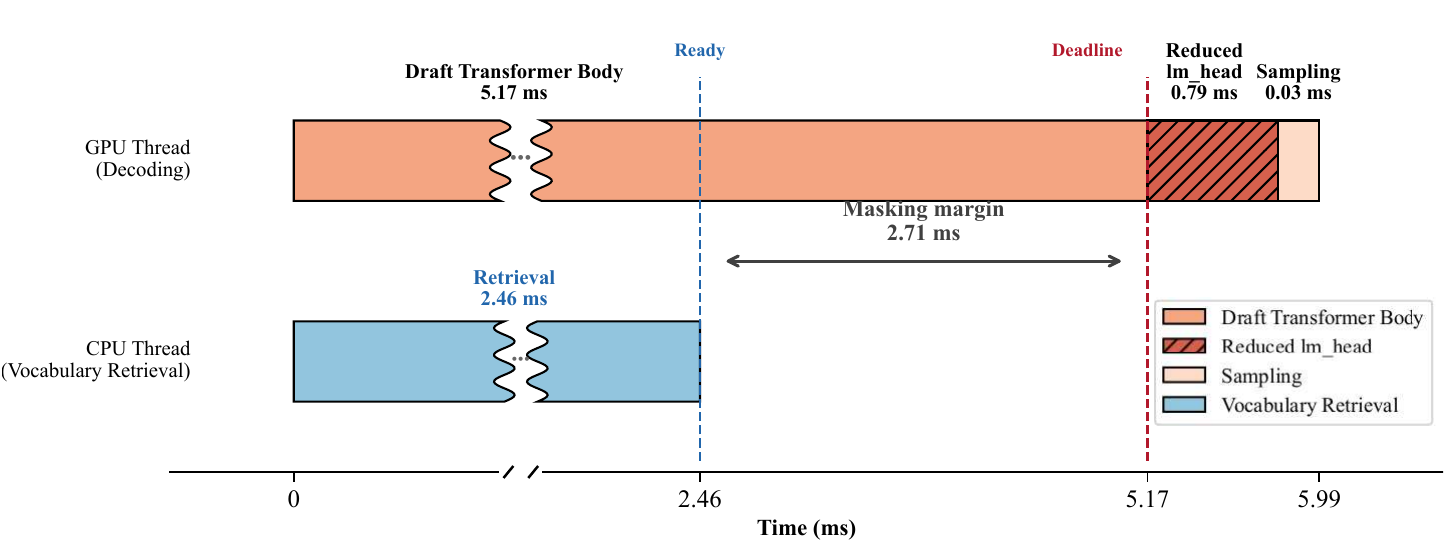}
    \caption{\textbf{Latency breakdown: CPU retrieval vs. GPU decoding.} Vocabulary retrieval is ready after 2.46 ms, leaving a 2.71 ms margin before the 5.17 ms draft Transformer body reaches the reduced-head deadline. The subsequent reduced LM head and sampling take 0.79 ms and 0.03 ms, respectively.}
    \label{fig:latency_breakdown}
\end{figure}

Vocabulary retrieval, including HNSW search and graph traversal, runs asynchronously on the CPU while the GPU executes the next draft Transformer body. Figure~\ref{fig:latency_breakdown} shows a retrieval-triggered step in which both paths start at time zero. The retrieval result becomes ready after 2.46 ms, whereas the draft Transformer body reaches the reduced LM-head boundary at 5.17 ms. Because the retrieved token IDs are needed only when forming the active output projection, this boundary is the deadline for using them in the current draft launch. The measured execution therefore provides a 2.71 ms masking margin.

After the deadline, the reduced LM head takes 0.79 ms and sampling takes 0.03 ms, bringing the illustrated draft step to 5.99 ms in total. Retrieval is consequently outside the critical path in this configuration. Full masking is not required for non-blocking execution: if a retrieval result misses the current reduced-head deadline, decoding proceeds with the existing active vocabulary and merges the returned candidates into a subsequent launch. A late result may postpone a small number of vocabulary hits, but it does not introduce a synchronization stall or alter target-model verification. The HNSW index is constructed once offline and incurs no online construction cost.

\section{Qualitative Analysis of Dynamic Prefetching}

\begin{figure}[H]
    \centering
    \includegraphics[width=0.72\textwidth]{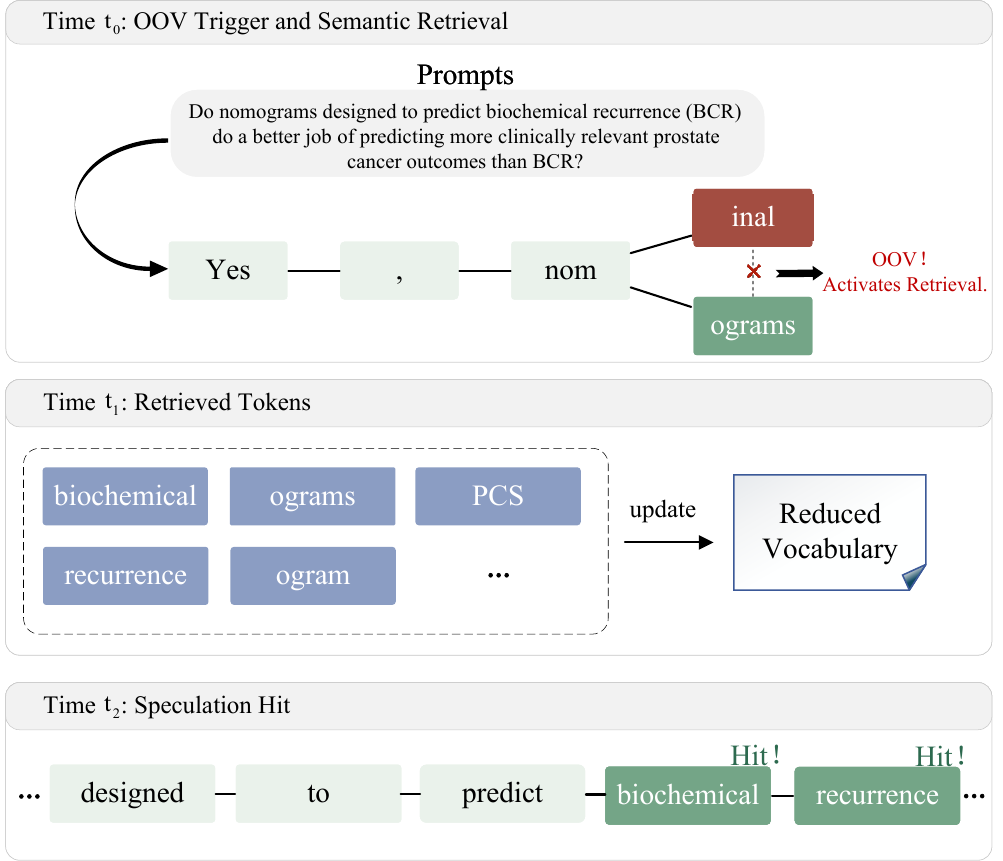}
    \caption{\textbf{Dynamic prefetching workflow.} An OOV event at ``\textit{nomograms}'' triggers retrieval of domain-relevant terms that support a later draft prediction.}
    \label{fig:process_flow}
\end{figure}

Figure~\ref{fig:process_flow} illustrates a retrieval event in a prostate-cancer prognosis trace. At time $t_0$, the static draft vocabulary fails to predict the suffix of ``\textit{nomograms},'' triggering semantic retrieval. At $t_1$, the semantic and graph indices retrieve related terms such as ``\texttt{biochemical},'' ``\texttt{recurrence},'' and ``\texttt{PCS},'' which are inserted into the dynamic buffer. At $t_2$, these prefetched tokens allow the draft model to propose ``\textit{biochemical recurrence}.'' The example illustrates how a reactive OOV event can prefetch the semantic neighborhood of an emerging topic and amortize retrieval over subsequent generation.

\section{Sampling Robustness}

We evaluate temperature-only sampling on the specialized-domain setting at $T=0.7$ and $T=1.0$, without an additional top-$p$ truncation variable. Table~\ref{tab:sampling} reports MAL and speedup normalized to FR-Spec at the same temperature.

\begin{table}[H]
\centering
\small
\setlength{\tabcolsep}{5pt}
\renewcommand{\arraystretch}{1.08}
\begin{tabular}{@{}lcccc@{}}
\toprule
\textbf{Method}
& \shortstack{\textbf{MAL}\\$T=0.7$}
& \shortstack{\textbf{Speedup}\\$T=0.7$}
& \shortstack{\textbf{MAL}\\$T=1.0$}
& \shortstack{\textbf{Speedup}\\$T=1.0$} \\
\midrule
FR-Spec (Static 32k) & 3.12 & 1.00$\times$ & 2.68 & 1.00$\times$ \\
SpecVocab            & 3.26 & 1.05$\times$ & 2.82 & 1.04$\times$ \\
\rowcolor{gray!10}
EvoSpec              & \textbf{3.31} & \textbf{1.08$\times$} & \textbf{2.87} & \textbf{1.06$\times$} \\
\bottomrule
\end{tabular}
\caption{Sampling robustness on specialized-domain averages. Speedup is normalized to FR-Spec under the same sampling temperature.}
\label{tab:sampling}
\end{table}

As the temperature increases, the target distribution broadens and the relative acceleration gain narrows. EvoSpec nevertheless retains the highest MAL and speedup at both temperatures.

\paragraph{Assets and licenses.}

All models and datasets are existing public assets. Qwen3 checkpoints are used under Apache-2.0, Llama-3.2 checkpoints under the Llama 3.2 Community License, SlimPajama under Apache-2.0 and its documented source-subset terms, HumanEval and PubMedQA under MIT, Pile of Law under CC BY-NC-SA 4.0 and its upstream-source terms, and LMSYS-Chat-1M under its dataset-specific license agreement. Spec-Bench and its constituent benchmarks are used under their respective original licenses.

\end{document}